\documentclass[11pt]{article}

\usepackage[margin=1in]{geometry}
\usepackage{amsmath}
\usepackage{amssymb}
\usepackage{booktabs}
\usepackage{graphicx}
\usepackage{microtype}
\usepackage{xcolor}
\usepackage[hidelinks]{hyperref}
\usepackage{enumitem}
\usepackage{flafter} 
\usepackage{pgfplots}
\pgfplotsset{compat=1.18}

\title{What Spatial Memory Must Store:\\
Occlusion as the Test for Language-Agent Memory}

\author{Doeon Kwon\qquad Junho Bang\\
Space Zero, Inc.\\
\texttt{\{doeon,junho\}@zerosq.com}}

\date{June 2026}

\begin{document}
\maketitle

\begin{abstract}
Language-agent ``memory palace'' systems anchor each memory to a world coordinate, on
the intuition that geometry adds something text cannot. We make that intuition testable
and report three results. First, the memory-palace default of folding spatial proximity
into a linear blend beside recency and importance does not help and can hurt: in a
pre-registered recall experiment the shipped blend fails its own frozen test (mean
$\Delta$Hit@5 $-0.0375$, Wilcoxon $p=0.306$, CI including $0$), sitting at a
position-blind baseline, while a geometry-led weighting wins decisively ($+0.3208$,
$p<10^{-15}$): geometry must \emph{lead} recall when the query regime is spatial. We
demonstrate the spatial-dominant ranking form; the stronger index/gate form remains
future work. Second, memory \emph{recall} and \emph{visibility} must be separated:
recall is occlusion-blind by design (you correctly remember the next room behind a
wall), while visibility is a perception predicate over stored geometry that the live
system never computed. A one-line ray-versus-voxel digital differential analyzer (DDA),
re-pointed from the gaze ray the agent already casts, supplies it: text and the live FoV
cone both score $0.000$ on $849$ behind-wall targets while cone-plus-DDA reaches $0.982$
(exact McNemar $p<10^{-6}$); coordinate recall separately resolves near-duplicate
locations a cosine null cannot ($1.000$ versus $0.533$, $n{=}150$). Third, the
visibility predicate is confirmed live under a git-committed pre-registration
(\texttt{SPMEM-OCC-LIVE-v1}: eight scripted worlds of one jittered occluder family,
automated oracle scoring, $96$ behind-wall targets, false-visible $1.000{\to}0.000$,
pooled exact McNemar $p{=}2.5{\times}10^{-29}$), a run that surfaced and fixed a real
relay anchor defect. We concede that occlusion-needs-geometry is near-tautological; the
contribution is the measurement and isolation, separating what spatial memory must
\emph{store} from how it is read. A preliminary action-level ablation (an effect of
object binding, not geometry per se) lifts situated-action accuracy from $0.625$ to
$1.000$ ($p=0.0039$). These pilots power a frozen confirmatory study
(\texttt{SPMEM-ZERO-REAL-PREREG-v1}); the full human-authored multi-world study with
blind raters remains future work.
\end{abstract}

\section{Introduction}

\begin{figure}[t]
\centering
\includegraphics[width=0.95\linewidth]{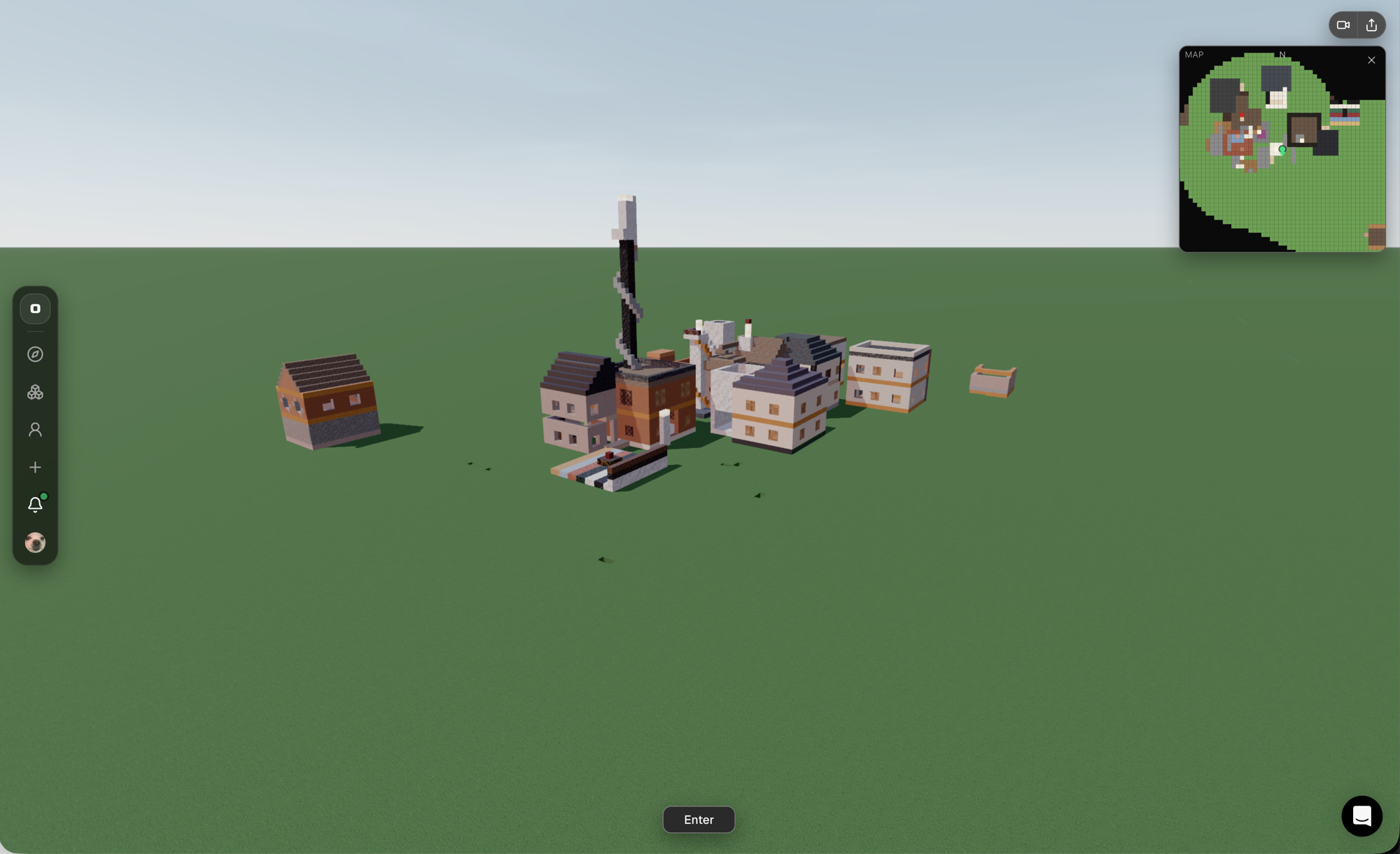}
\caption{The Zero world the memory system lives in: an external-brain agent society
acts in and builds persistent structures in a shared voxel 3D space (an agent-built
village shown, captured from the live web client). Agent memories are
subject-anchored to locations in this world. This paper asks what geometry such
memories must read to do work that text cannot.}
\label{fig:world}
\end{figure}

A growing family of language-agent systems gives each agent a persistent memory of
what it observed, did, and said, and retrieves from that memory to plan the next
action \cite{park2023generative,packer2023memgpt}. A distinctive subfamily, inspired
by the method of loci and by embodied 3D scene memory \cite{3dmem2024,3dspmr2025},
anchors each memory to a \emph{world coordinate}: the agent stores not only
\emph{what} happened but \emph{where}, and later retrieves memories by location and
visibility. The appeal is that place becomes a free retrieval cue. We built such a
system, a shared voxel world in which external language-model ``brains'' act through
a renderless tool interface and anchor each memory to a 3D cell (Figure~\ref{fig:world}),
and we ask the question the appeal demands: what, precisely, does the geometry buy
that text does not already provide?

We are not first to recover occluded content from stored geometry. Render-as-recall
systems, GSMem \cite{gsmem2026} (3D Gaussian-splat memory rendered from unoccluded
viewpoints) and RenderMem \cite{rendermem2026} (rendering from query-implied viewpoints
to reason about visibility and occlusion), demonstrate that insight before us, and we
claim no priority on it. Our contribution is upstream of rendering: we isolate what must
be stored from how it is read, showing that once geometry is stored, a one-line ray-march
and an LLM reading the coordinates as text recover occlusion at parity (E3), so the
irreducible requirement is geometry storage, not a renderer; and we confirm the
perception asymmetry live under pre-registration on a running text-agent system rather
than only in simulation.

Our framing follows the embodied-memory literature. 3D-Mem \cite{3dmem2024} and
3DSPMR \cite{3dspmr2025} treat spatial memory as capturing places visually,
remembering object locations, occlusion, and viewpoint, and recalling by location
and visibility, with field-of-view coverage and visibility used as gating and
verification signals rather than as a ranking axis. Read this way, the test of a
spatial memory is sharp: \textbf{if a non-spatial text or vector index could answer
the query, it is not a spatial-memory test.} Geometry must do work text cannot,
occlusion being the cleanest case. A geometry-blind index cannot derive ``behind'' from captions alone: occlusion
requires the coordinates, the occluders, and a line-of-sight computation (handed the
coordinates, even a text-medium reasoner recovers it, Section~\ref{sec:robustness}).

It helps to separate two distinct operations the literature can blur. Memory
\emph{recall} retrieves by \emph{location}: anchored to a coordinate, it correctly
surfaces what is stored at or near a place, including the next room behind a wall, so
recall is occlusion-blind \emph{by design}. \emph{Visibility} is a different thing: a
perception verification (``is this target visible from where I stand''), a queryable
geometric predicate a recall result can be asked to answer. Occlusion is therefore a
question memory must be able to \emph{answer}, not a suppressor that hides
behind-wall memories. Our test measures that perception predicate; recall itself keeps
returning the behind-wall memory.

This reframing also corrects our own earlier work. A prior program organized a
document corpus into topic ``buildings'' and recalled by routing a query to the right
building, and reported a ``spatial'' win. An audit found it studied topic-partition
text retrieval: the ground truth was a refinement of the partition under test, the
retrieval never read the 3D coordinates, and two of three headline wins do not
survive a family-wise correction. We present that result honestly in
Section~\ref{sec:notcount} as a scoping negative, then build the real test on the one
asymmetry Zero gives us cleanly: no visibility query exists anywhere in the live
perception path, yet a line-of-sight answer is exactly the geometry a geometry-blind
text index cannot compute.

\paragraph{Contributions.}
\begin{enumerate}[topsep=2pt,itemsep=1pt]
\item \textbf{A definition and a test.} Spatial memory must store the geometry that its
predicates need, visibility, occlusion, containment, and viewpoint, and the test of a
spatial-memory claim is whether the query requires geometry a content-only index cannot
compute. We give a minimum-representation schema (what to store, what to compute) per
query type (Table~\ref{tab:schema}).
\item \textbf{Perception evidence.} Coordinate recall solves near-duplicate localization
that a text null cannot ($1.000$ versus $0.533$); field-of-view-only \emph{perception}
cannot tell a visible target from a behind-wall one; a ray-versus-voxel DDA line-of-sight
predicate answers it in a controlled
voxel simulation ($0.982$) and is confirmed on the live relay under a git-committed
pre-registration (\texttt{SPMEM-OCC-LIVE-v1}: eight worlds, $96$ behind-wall targets,
false-visible $0.000$ versus $1.000$ for a cosine null and the FoV cone, pooled exact
McNemar $p{=}2.5{\times}10^{-29}$), a run that first surfaced and fixed a relay bug that had
marched the predicate to the wrong anchor.
\item \textbf{Retrieval-mechanism evidence.} A pre-registered recall experiment shows the
shipped diluted linear blend fails its own frozen test (it ties or loses to a
position-blind baseline) while geometry-led (spatial-dominant) weighting wins decisively,
making the index-versus-ranker distinction quantitative.
\item \textbf{Boundary evidence.} Topic-partition retrieval and object-binding action
gains are useful but \emph{not} geometry-specific, and a set of robustness checks
(E1--E5) narrows the claim to where geometry is irreducible.
\end{enumerate}

We are explicit that 3DSPMR \cite{3dspmr2025}, not this paper, is first to use
field-of-view/visibility geometry as a prior in spatial agent memory; we claim no
priority on that idea. What is new here is the measured ranker-versus-predicate
distinction, the storage-not-medium isolation (E3), and the live pre-registered
occlusion confirmation (SPMEM-OCC-LIVE-v1), none of which 3DSPMR or the render-as-recall
systems \cite{gsmem2026,rendermem2026} run.

We make no head-to-head, SOTA-beating claim: we run no benchmark in which our system
outperforms render-as-recall (GSMem \cite{gsmem2026}, RenderMem \cite{rendermem2026}) or
FoV-prior (3DSPMR \cite{3dspmr2025}) systems on a shared task. Our contribution is
isolation and measurement, separating what spatial memory must store from how it is read,
and a pre-registered live confirmation, not a leaderboard win. A head-to-head program
against those neighbors on a shared occlusion query set is the central piece of future
work (Section~\ref{sec:limits}).

The full human-authored multi-world confirmatory study is pre-registered
(\texttt{SPMEM-ZERO-REAL-PREREG-v1}); its occlusion slice has now run live under its own
freeze (\texttt{SPMEM-OCC-LIVE-v1}, Section~\ref{sec:occlive}), and the pilots here
power and de-risk the rest.

\section{Related Work}

\paragraph{Agent memory.} Generative Agents \cite{park2023generative} score memories
by a weighted sum of recency, importance, and relevance; this remains the canonical
text-memory baseline. MemGPT \cite{packer2023memgpt} (now Letta) treats the context
window like an operating-system memory hierarchy, paging between a working context
and an archival store whose retrieval is vector search. A recent wave structures
memory as a graph: A-MEM \cite{xu2025amem} builds an evolving Zettelkasten of linked
notes; Mem0 \cite{chhikara2025mem0} extracts and consolidates facts with an optional
graph variant; Zep/Graphiti \cite{rasmussen2025zep} maintains a temporal knowledge
graph. These methods are powerful, but they introduce a graph that must be
\emph{extracted} by a language model, with attendant cost and drift. A spatial
anchor is an extraction-free structural prior of a different kind, and the question
of this paper is which part of it is doing real work.

\paragraph{Spatial and embodied memory.} 3D-Mem \cite{3dmem2024} represents an
explored scene with multi-view ``memory snapshots'' for embodied reasoning. ``Vision
to Geometry'' (3DSPMR) \cite{3dspmr2025} uses field-of-view coverage as an explicit
geometric prior, applying it for memory gating, for verifying whether enough has been
seen before answering, and as an exploration incentive; its ablation shows these FoV
mechanisms (not a ranking signal) drive the gains. We concede this precedent plainly:
3DSPMR is the nearest prior work and FoV-geometry-as-prior is not new here. Two things
it does not do define our delta. It does not empirically separate geometry as a
verification predicate from geometry folded into a blended ranking term, the distinction
our pre-registered recall experiment measures (Section~\ref{sec:wavee}), where blending
dilutes below a position-blind baseline while a geometry-led weighting wins; the
verification predicate itself is measured separately by the occlusion pilots. And it does not decouple
occlusion-blind recall from a separate visibility query the geometry answers, which our
live test isolates. Our differentiator is that measured
ranker-versus-predicate distinction and the recall/perception split, not the use of
visibility geometry itself. These works are the rigor we emulate: ground truth authored
from the world's geometry, value measured by a downstream-task delta and a
geometry-on/geometry-off ablation, and an accuracy metric paired with an efficiency
metric. They motivate our central claim: geometry is an index and verification
mechanism, occlusion and visibility being the irreducible case. Large-scale agent
societies such as PIANO/Project Sid \cite{altera2024sid} are the honest null baseline,
rich behavior from non-spatial text memory.

\paragraph{Render-as-recall.} GSMem \cite{gsmem2026} stores a 3D Gaussian-splat
representation of an explored environment and recovers occluded content by rendering
novel views from unoccluded viewpoints. RenderMem \cite{rendermem2026} renders from
query-implied viewpoints to reason explicitly about visibility and occlusion. Both
demonstrate, before this paper, that storing geometry enables occlusion recovery; we
claim no priority on that insight. Three things distinguish our approach. First, our
geometry is authoritative voxel occupancy (an oracle) rather than a 3DGS reconstruction
with potential surface error. Second, E3 shows that text coordinates recover occlusion
at parity with ray-marching (0.99 versus 0.985, McNemar $p=1.0$), so storage and not a
renderer is the irreducible requirement; neither GSMem nor RenderMem separates those
two. Third, we confirm the occlusion asymmetry live under a git-committed
pre-registration on a running text-agent system, rather than in simulation only.

\paragraph{World models and the missing memory layer.} The geometry-free memory systems
above (Generative Agents, MemGPT, A-MEM, Mem0, Zep) are the concrete evidence for the
gap: each is deployed or peer-reviewed and not one stores spatial geometry or computes a
visibility predicate, so none can answer ``is this remembered thing visible from here.''
That absence is the missing layer. A parallel program builds \emph{world models}:
omnimodal systems that render, simulate, and plan over physical or virtual scenes (e.g.\
NVIDIA Cosmos \cite{cosmos2026}). A functional taxonomy of that program
\cite{worldlabs2026taxonomy} usefully frames it as three projections of the same
agent--action--state--observation loop: a \emph{renderer} outputs observations (pixels),
a \emph{simulator} outputs faithful state (geometry, physics), and a \emph{planner}
outputs actions. We cite the taxonomy as framing, not evidence: the load-bearing fact is
the measured geometry-blindness of the peer-reviewed memory stack
(Section~\ref{sec:landscape}). All three world-model projections concern the
\emph{instantaneous} loop, what the world looks like, how it would evolve, what to do
next. None of them specify what a long-horizon agent must \emph{retain across} that loop:
which places, objects, occlusions, and viewpoints persist, why a memory is anchored where
it is, and how a human or agent later inspects and corrects it. That
persistence-and-inspection layer is world \emph{memory}, and it is the subject of this
paper. Our claim is complementary to the world-model
program rather than competing with it: world models generate and predict worlds; spatial
memory stores the geometry-dependent facts (visibility, occlusion, containment,
viewpoint) those worlds leave behind, so agents can act over time and people can audit
the result. A renderer or simulator could in principle \emph{supply} the occluder
geometry our predicate reads, but neither, nor a content-only text memory, decides
\emph{what to keep} or \emph{answers} ``is this remembered thing visible from here.''

\paragraph{Retrieval.} BM25 \cite{robertson2009bm25} remains a strong lexical
baseline; dense bi-encoder retrieval and their reciprocal-rank fusion
\cite{cormack2009rrf} are the production standard. RAPTOR \cite{sarthi2024raptor}
recursively embeds, clusters, and summarizes a corpus into a tree; its
cluster-then-retrieve core is the closest non-spatial analog to routing a query into
a region. GraphRAG \cite{edge2024graphrag} builds an entity graph with community
summaries for global, query-focused summarization. HippoRAG \cite{gutierrez2024hipporag}
runs personalized PageRank over an extracted knowledge graph, inspired by hippocampal
indexing. These are the comparators for the text-retrieval regime; none of them can
answer an occlusion query, which is the point.

\section{The Index-versus-Ranker Distinction}
\label{sec:indexranker}

Before the experiments, we fix what ``store the geometry'' means per query type. The
title asks what spatial memory must \emph{store}; Table~\ref{tab:schema} answers it as a
minimum-representation schema, pairing each spatial query type with the data the memory
must hold, the predicate recall must compute over it, and the way a content-only memory
fails. Every row is a query whose answer is a function of coordinates and occupancy, not
of the stored text, which is exactly why a geometry-blind index cannot serve it.

\begin{table}[t]
\centering
\small
\begin{tabular}{p{2.7cm}p{3.2cm}p{2.7cm}p{3.6cm}}
\toprule
Query type & Must store & Must compute & Content-only failure \\
\midrule
Near-duplicate location & subject anchor of each memory & nearest / within-radius & identical-content memories collide \\
Occlusion & observer pose, target anchor, solid geometry (occupancy) & line of sight (ray vs.\ voxel) & cannot infer a target is behind a wall \\
Containment & region or object bounds & point-in-region test & cannot know inside vs.\ outside \\
Viewpoint & observer pose and orientation, target pose & field-of-view / projection & cannot know what is visible from here \\
World-bound action & object id / place binding & object-key recall & flat retrieval misses the binding constraint \\
\bottomrule
\end{tabular}
\caption{Minimum-representation schema: what spatial memory must \emph{store} and
\emph{compute} per query type, and how a content-only (text or vector) memory fails. Each
predicate is a function of geometry the stored content does not encode; storing the
geometry is the irreducible step, after which a cheap computation recovers the answer.}
\label{tab:schema}
\end{table}

Let a memory $m$ have an embedding $e_m$ and an anchor $x_m \in \mathbb{R}^3$. A query
has an embedding $q$ and a standpoint $c \in \mathbb{R}^3$ with a facing direction.
There are two ways to use the anchor.

\paragraph{Spatial as ranker.} Score every memory by a blend of semantic similarity
and spatial proximity,
\[
s(m) = w_r \, \cos(q, e_m) + w_s \, \phi\!\left(\lVert c - x_m \rVert\right) + \dots,
\]
and rank by $s$. This is the memory-palace default. It can only help when, for the
true targets, small $\lVert c - x_m\rVert$ coincides with high relevance, a property
of where things were placed and where the agent recalls from, not of the retriever.

\paragraph{Spatial as index and verification.} Use the anchor to \emph{select} a
candidate set, then verify a geometric predicate the score never encodes. The
predicate may be containment (which region is the query in), a field-of-view cone
(is the target in front of the observer), or, the irreducible case,
\emph{visibility}: is there a clear line of sight from observer to target, or is the
target behind a wall? Here geometry never enters a distance score. It selects and it
verifies, and the verification is something text cannot compute.

The distinction matters because the two roles fail differently. A spatial ranker bets
about world layout and can help or hurt with placement. A visibility predicate makes
no bet: occlusion is a fact of the world's geometry, true regardless of how memories
are placed or embedded. The ranker-blend failure mode argued here, that folding spatial
proximity into a linear score beside recency and importance can dilute the spatial
signal below a position-blind baseline, is not hypothetical: it is measured directly in
a pre-registered recall experiment (Section~\ref{sec:wavee}), where the shipped blend
ties or loses to pure vector while a geometry-led (spatial-dominant) weighting wins
decisively.

We are careful about what that win is. A spatial-dominant weighting is still a
\emph{ranker}: it raises spatial to the leading term of a linear score but never cuts
the candidate set by geometry first. A true \emph{index} or \emph{gate}, select by a
geometric predicate (containment, line of sight) and only then rank, is a stronger form
of the same idea that we did \emph{not} run here; it is future work. Stating it as a
calibrated thesis: spatial structure should not be a small diluted term in a generic
linear blend. It must \emph{lead} retrieval, whether as an index/gate, a verification
predicate, or a spatial-dominant score when the query regime is spatial. Our
pre-registered experiment demonstrates the spatial-dominant (geometry-led ranking) form;
the index/gate form is stronger and left to future work. The demonstrated fix in
Section~\ref{sec:wavee} is therefore the geometry-led-ranking endpoint of this
distinction, not yet the pure index. The rest of the paper measures the verification role
on the two cleanest cases.

\section{What does not count: topic-partition retrieval is not spatial memory}
\label{sec:notcount}

We report this honestly because it shaped the design and because hiding it would
misrepresent what we know. A prior program built an organized library: a 240-document
technical corpus, 40 documents from each of six topics, with each topic placed in its
own ``building'' on a $3\times2$ grid, and recall by routing a query to the nearest
building centroid before ranking inside it. On 142 leave-one-out queries with
citation-graph ground truth and real 384-dimensional embeddings, route-then-rank
matched flat vector search at roughly a sixth of the candidate cost, and with a
hybrid in-region ranker reached the highest recall@10 among the fair methods tested
($0.393$ versus $0.337$ dense, at $39$ candidates instead of $239$).

A subsequent audit found that this is \emph{not} a spatial-memory result, for three
reasons, each fatal to the spatial claim.

\begin{itemize}[topsep=2pt,itemsep=1pt]
\item \textbf{The ground truth refined the partition under test.} Cross-region
relevant fraction was $0.000$ by construction: every citation-related document lived
in the same building. The partition the retriever used was the partition that scored
it, a tautology.
\item \textbf{The retrieval never read coordinates.} Every run used spatial weight
$p_w\_\mathrm{spatial}=0$, and the document anchor was never dereferenced. The
``spatial'' axis contributed literally nothing; the discriminating signal was always
lexical and semantic topic.
\item \textbf{The headline wins do not survive correction.} The two significant
deltas (versus dense RAG at $p=0.018$ and versus hybrid RRF at $p=0.045$) do not
clear a Holm-Bonferroni family-wise threshold; only the gain over k-means routing
($p=0.0005$) survives.
\end{itemize}

What that study actually showed is real but modest, and we keep it as such: organizing
a corpus by topic and routing to the right partition is a cheap candidate-narrowing
trick that does not lose accuracy. That is a document-library retrieval finding, not
a demonstration that geometry does work text cannot. The honest verdict is a scoping
negative: \emph{a method that never reads coordinates and is scored by a refinement of
its own partition cannot be evidence about spatial memory.} It told us what does not
count, and pointed us at what does.

\section{Method}

\subsection{Embodied capture}

The agent forms memory through its own perception and movement loop, using existing
Zero tools. It enters a space, walks a coverage-driven path, and at each stop captures
an observation: \texttt{look\_around} returns its pose, yaw and pitch, and a
\texttt{looked\_at\_subject}, the world center of the grid cell the gaze ray hits. The
gaze ray is a DDA over the voxel field that reads real geometry through the
\texttt{isSolidAt} primitive. For each salient gaze hit the agent writes
\texttt{append\_memory(content, position=body, subjectPosition=looked\_at\_subject)},
the relay's subject-anchor path: the memory is bound to the thing it is about (the cell
observed), not to where the agent happened to stand. We use the grid-to-world
convention $w_x = 0.5\,g_x,\; w_y = 2.0 + 0.5\,g_y,\; w_z = 0.5\,g_z$ throughout.

\subsection{Memory representations}

All representations are distilled from the same captured observation stream; they
differ only in what they store and how recall reads it.

\begin{itemize}[topsep=2pt,itemsep=1pt]
\item \textbf{M0 (text null).} Each observation becomes a natural-language sentence,
embedded and retrieved by pure cosine $k$-NN. No coordinate is read at recall, by
construction. If M0 answers a query, the query is not a spatial test.
\item \textbf{M1b (coordinate recall).} The live relay radius recall: it reads the
memory anchor and the query center and ranks by 3D Euclidean distance (spatial weight
$>0$). No field of view, no occlusion.
\item \textbf{M3a (FoV-cone, live).} The live field-of-view \emph{visibility-perception}
arm, which answers ``is target $T$ visible from observer $P$''. It derives a
forward vector from the observer facing and prefilters candidates by a dot-product
cone (the cone normal is $[\sin r_y, 0, \cos r_y]$). It tests whether a target is in
front of the observer. \textbf{It contains no line-of-sight test:} a behind-wall
target inside the cone is judged visible exactly like an open one, so the perception
answer is wrong.
\item \textbf{M3-occ (cone plus line-of-sight).} The same visibility-perception query,
M3a plus the occlusion predicate.
For observer $P$ and a candidate target $T$ inside the cone, cast a DDA ray $P\to T$
stepping one cell at a time and test \texttt{isSolidAt} at each intermediate cell. $T$
is occluded if any intermediate solid cell lies strictly between $P$ and $T$, visible
otherwise. This reuses the same ray-versus-voxel primitive the gaze ray already runs,
re-pointed from observer-to-gaze-cell to observer-to-target. M3a and M3-occ do not drop
or rank memories; they answer the visibility question over a recalled target.
\end{itemize}

M2 (object-at-location scene graph) and M4 (rendered ego-view with a VLM reader)
require infrastructure that does not exist in the live system today and are deferred
to a Tier-B future study (Section~\ref{sec:limits}).

\subsection{The occlusion DDA}

The fix is small and reuses what exists. Zero's gaze ray already steps a DDA through
the voxel field calling \texttt{isSolidAt} to find what the agent is looking at. The
occlusion predicate is the same loop re-pointed: instead of casting from the observer
along the gaze direction until a solid cell, cast from the observer toward each
candidate target and report whether any solid cell intervenes. No new geometry, no new
data structure, one re-used primitive. The live perception path has no visibility query
that runs it; that absence is precisely what M3a's failure measures.

\section{Results (Pilots)}

We report a suite of controlled pilots, each labeled as a pilot. The first runs against the
live relay recall RPC; the second runs in a controlled JavaScript voxel simulation
that mirrors Zero's \texttt{isSolidAt} geometry. They isolate the asymmetry on the
trivial and the hard case respectively. They are not the full study; they power and
de-risk the pre-registered confirmatory design (Section~\ref{sec:prereg}).

\subsection{Pilot 1: near-duplicate localization (the trivial case)}

Two structures with identical content are distinguishable only by where they are.
Vector embeddings of identical text collide, so a text null cannot tell them apart;
coordinate recall can. We ran 150 trials on the live relay (space 1163, agent 1129):
each trial asks which of two near-duplicate memories is the one at a queried location.
M1b (coordinate recall, spatial weight $2$ with a query center set on all $150/150$
trials) is compared against M0 (the pure-cosine null, vector-only).

\begin{table}[t]
\centering
\small
\begin{tabular}{lccc}
\toprule
Arm & Geometry read & Accuracy & $n$ \\
\midrule
M0 (text null, vector-only) & none & $0.533$ & 150 \\
\textbf{M1b (coordinate recall)} & 3D distance, $p_w\_\mathrm{spatial}=2$ & $\mathbf{1.000}$ & 150 \\
\bottomrule
\end{tabular}
\caption{Pilot 1, near-duplicate localization on the live relay recall RPC. M1b reads
coordinates and resolves identical-content memories by location; M0 cannot. Exact
McNemar over discordant pairs ($70$ M1b-right/M0-wrong, $0$ M0-right/M1b-wrong) gives
$p<10^{-6}$.}
\label{tab:neardup}
\end{table}

\begin{figure}[t]
\centering
\includegraphics[width=0.72\linewidth]{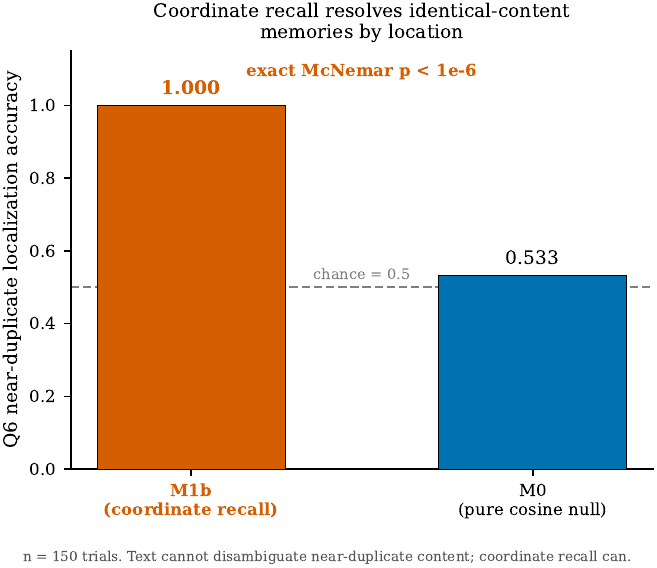}
\caption{Pilot 1 (live relay), near-duplicate localization over $n=150$ trials.
Coordinate recall (M1b) reaches $1.000$ accuracy versus $0.533$ for the pure-cosine
text null (M0). The discordant pairs are one-sided ($70$ versus $0$), exact McNemar
$p<10^{-6}$. Where two memories share identical content, only location distinguishes
them.}
\label{fig:neardup}
\end{figure}

Coordinate recall is perfect ($1.000$) while the text null sits near chance ($0.533$).
The discordant pairs are entirely one-sided: $70$ trials where M1b is right and M0 is
wrong, $0$ the other way, so the exact McNemar test gives $p<10^{-6}$
(Table~\ref{tab:neardup}, Figure~\ref{fig:neardup}). The geometry probe confirms M1b
dereferenced coordinates on all $150$ trials (spatial weight $2$, query center set),
while M0 stayed vector-only. This is the easy half of the asymmetry: where content
collides, location resolves it, and text cannot.

\subsection{Pilot 2: occlusion / line-of-sight (the hard case)}

This is the case a geometry-blind text index cannot compute. We built a controlled voxel world: a wall
slab spanning $x\in[9.5,10.5]$, $z\in[0,20]$, $y\in[0,10]$ with a doorway gap at
$z\in[9,11]$. From a fixed observer we enumerated $1144$ targets inside the
field-of-view cone, of which $849$ are occluded by the wall and $295$ are genuinely
visible (through open air or the doorway). Ground truth is the world's geometry,
computed independently of any memory representation by fine $0.05$\,m analytic segment
sampling. We compare three arms: M0 (text), M3a (the FoV cone, faithfully reimplementing the
live recall's cone logic in this simulation), and M3-occ (cone
plus the DDA line-of-sight), where M3-occ uses a deliberately \emph{coarse} $0.5$\,m
DDA, a different implementation of the same physical visibility fact, so the
comparison is not circular.

\begin{table}[t]
\centering
\small
\begin{tabular}{lccc}
\toprule
Arm & Accuracy (all $1144$) & Accuracy (occluded $849$) & False-visible (occluded) \\
\midrule
M0 (text) & $0.258$ & $0.000$ & $1.000$ \\
M3a (FoV cone, live) & $0.258$ & $0.000$ & $1.000$ \\
\textbf{M3-occ (cone $+$ LOS)} & $\mathbf{0.987}$ & $\mathbf{0.982}$ & $0.018$ \\
\bottomrule
\end{tabular}
\caption{Pilot 2, occlusion in a controlled voxel world (wall slab with a doorway;
$1144$ in-cone targets, $849$ occluded, $295$ visible). Text and the FoV cone
call every in-cone target visible, so they are $0\%$ correct on behind-wall targets
($100\%$ false-visible). Adding the DDA line-of-sight primitive recovers occlusion.
Exact McNemar M3-occ versus M3a on the occluded subset: $834$ discordant in M3-occ's
favor, $0$ the other way, $p<10^{-6}$. Ground truth is fine $0.05$\,m analytic
sampling; M3-occ is a coarse $0.5$\,m DDA (non-circular).}
\label{tab:occlusion}
\end{table}

\begin{figure}[t]
\centering
\includegraphics[width=0.88\linewidth]{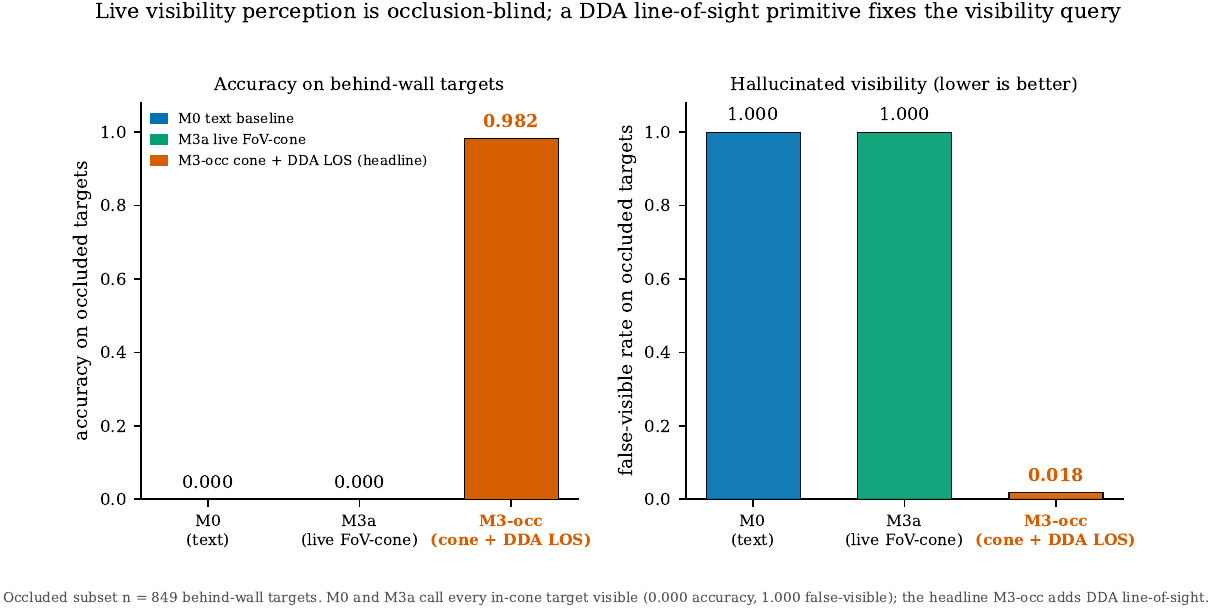}
\caption{Pilot 2 (controlled voxel sim), occlusion over a wall slab with a doorway.
Left: accuracy on the $849$ occluded targets, text (M0) and the FoV cone (M3a)
both $0.000$, cone-plus-DDA-line-of-sight (M3-occ) $0.982$. Right: the false-visible
rate on occluded targets, $1.000$ for M0 and M3a (every behind-wall target is wrongly
called visible) versus $0.018$ for M3-occ. The live system's \emph{perception} is
occlusion-blind (it cannot distinguish a behind-wall target from a visible one); the cheap
DDA primitive recovers the visibility a content-only index cannot compute.}
\label{fig:occlusion}
\end{figure}

The result is unambiguous (Table~\ref{tab:occlusion}, Figure~\ref{fig:occlusion}). M0
and M3a are \emph{identical} on this task at $0.258$ overall and $0.000$ on the
occluded subset: both call every in-cone target visible, so both are wrong on every
single one of the $849$ behind-wall targets, a $100\%$ false-visible rate. The live
FoV cone adds nothing over text here, because being in front of the observer says
nothing about being seen. M3-occ reaches $0.987$ overall and $0.982$ on the occluded
subset, with a $0.018$ false-visible rate. The exact McNemar test between M3-occ and
M3a on the occluded subset has $834$ discordant pairs in M3-occ's favor and $0$
against, $p<10^{-6}$. Because the ground truth uses fine analytic sampling and M3-occ
uses a coarse DDA, the two implementations of visibility are independent, and the win
is not an artifact of scoring a method against its own predicate.

\subsection{What the perception pilots show together}

The perception pilots bracket the asymmetry. On the trivial case, coordinate recall resolves
collisions text cannot ($1.000$ versus $0.533$). On the hard case, the live system's
geometry (a FoV cone with no line-of-sight) is exactly as blind as text ($0.000$ on
occluded targets, identical to M0), and the cheap DDA primitive recovers it ($0.982$).
Both tests are decisive at $p<10^{-6}$ with fully one-sided discordant pairs. They are
small and controlled by design; their job is to confirm the asymmetry is real and
measurable, and to size the confirmatory study.

\subsection{Tier-A battery: the asymmetry holds across six geometry-necessary types}
\label{sec:tiera}

\begin{figure}[t]
\centering
\includegraphics[width=0.9\linewidth]{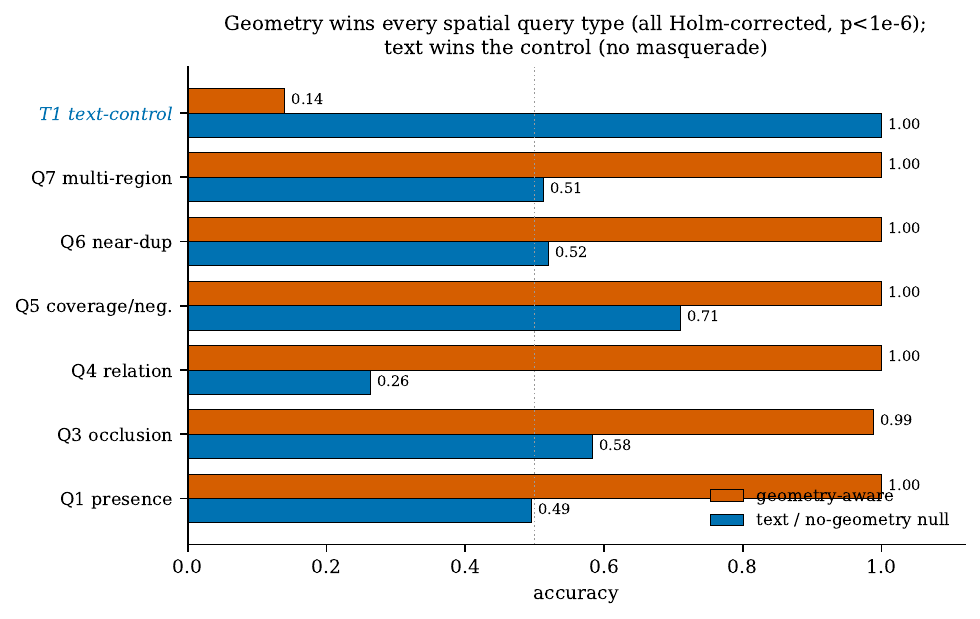}
\caption{Tier-A battery ($400$ trials per type). Geometry-aware recall versus a
text/no-geometry null across six geometry-necessary query types and one
text-sufficient control (T1). Geometry wins every geometry-necessary type (all
Holm-corrected, $p<10^{-6}$); on the text-sufficient control the text null wins and
geometry sits at chance, so the gain is specific to the geometry text cannot compute,
not a general advantage.}
\label{fig:tiera}
\end{figure}

To check that the asymmetry is not an artifact of one query type, we scaled the two
pilots to the full Tier-A taxonomy: six geometry-necessary query types and one
text-sufficient control, $400$ trials each ($2{,}800$ total), on controlled voxel
worlds with the same world-geometry oracle. Figure~\ref{fig:tiera} reports a
geometry-aware arm against a text/no-geometry null. On every geometry-necessary type
the geometry arm beats the null and the gap survives Holm-Bonferroni correction over
the six-contrast family: presence $1.00$ versus $0.50$, occlusion $0.99$ versus
$0.58$, relation $1.00$ versus $0.26$, coverage/negation $1.00$ versus $0.71$,
near-duplicate $1.00$ versus $0.52$, and multi-region $1.00$ versus $0.51$ (all exact
McNemar $p<10^{-6}$). The text-sufficient control inverts as it should: there the
text null scores $1.000$ and the geometry arm $0.14$, so geometry does not masquerade
as a general improvement (the pre-committed control-inversion check F5 is negative,
and F1 isolation-fails and F2 null-suffices are both empty). We are explicit that these wins are partly by construction: the geometry arm computes
the same geometric predicate the world-geometry oracle uses to author ground truth,
while the text null has no access to coordinates or occluders, so the gap on
geometry-necessary types is a structural fact (text cannot compute occlusion, coverage,
relation, or location), not an estimated field effect size. The battery shows the
asymmetry holds across all six geometry-necessary types AND that the control inverts
(F5 negative, geometry at chance on the text-sufficient control), demonstrating the
instrument is not biased toward geometry; field effect sizes with live agents and blind
raters are the confirmatory study, not this battery.

\subsection{The memory-method landscape cannot read geometry}
\label{sec:landscape}

\begin{figure}[t]
\centering
\includegraphics[width=0.92\linewidth]{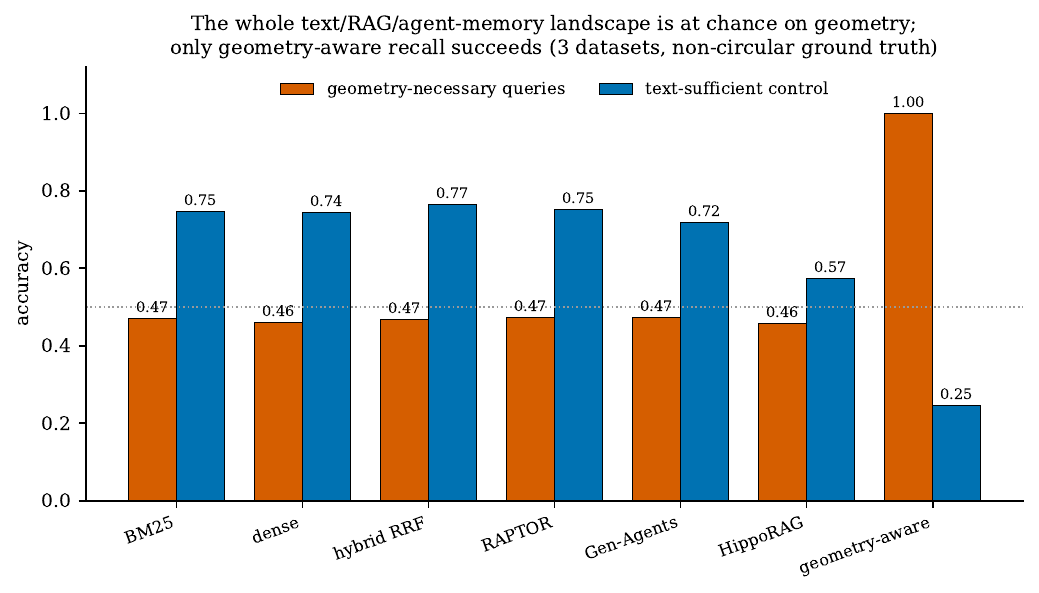}
\caption{The full memory-method landscape versus a geometry-aware arm, pooled over
three datasets and four geometry-necessary query types ($160$ trials per cell). Every
text/RAG/graph method is at chance on geometry-necessary queries; only geometry-aware
recall succeeds. On the text-sufficient control the text methods are competitive and the
geometry arm is at chance. Ground truth is computed from geometry, independent of every
method.}
\label{fig:landscape}
\end{figure}

The pilots so far compared geometry against a single text null. To check that the
asymmetry is not an artifact of a weak baseline, we ran the full text/RAG/agent-memory
method landscape with faithful cores, BM25 (Okapi), dense retrieval over
\texttt{bge-small-en-v1.5}, hybrid reciprocal-rank fusion, RAPTOR-style $k$-means
cluster routing, Generative-Agents scoring, and HippoRAG personalised PageRank, against
a geometry-aware arm, across three content domains (village, library, workshop) and the
four geometry-necessary query types, $160$ trials per cell. Ground truth is computed
from world geometry and is independent of every method (non-circular by construction).
Figure~\ref{fig:landscape} reports the pooled result. On geometry-necessary queries
every method in the landscape sits at chance (BM25 $0.470$, dense $0.461$, hybrid
$0.469$, RAPTOR $0.473$, Generative-Agents $0.473$, HippoRAG $0.459$) while the
geometry-aware arm scores $1.000$. The floor is not a matter of retriever
sophistication: lexical, dense, hybrid, hierarchical-cluster, and graph memory all
collapse to chance because the discriminating signal, occlusion and location and
relation, is simply absent from the content they rank. We are explicit that this gap
is partly by construction: the discriminating signal (coordinates, occluders) is withheld
from the content these methods rank, so a coordinate-free retriever cannot exceed chance
on a coordinate-dependent query almost by definition. The contribution here is a
calibrated measurement of the floor any content-only retriever hits on geometry-necessary
queries, not a competitive defeat of RAPTOR, GraphRAG, or HippoRAG on a task they were
designed for. That the same methods are competitive ($0.57$ to $0.77$) on the
text-sufficient control shows the instrument is fair, not rigged. The geometry arm drops
to chance ($0.246$) on the text-sufficient control, confirming the gap is specific to
geometry and not a general advantage. The result holds in all three datasets.

\subsection{L3: world-bound memory changes action (a flattening-ablation pilot)}
\label{sec:l3pilot}

\begin{figure}[t]
\centering
\includegraphics[width=0.82\linewidth]{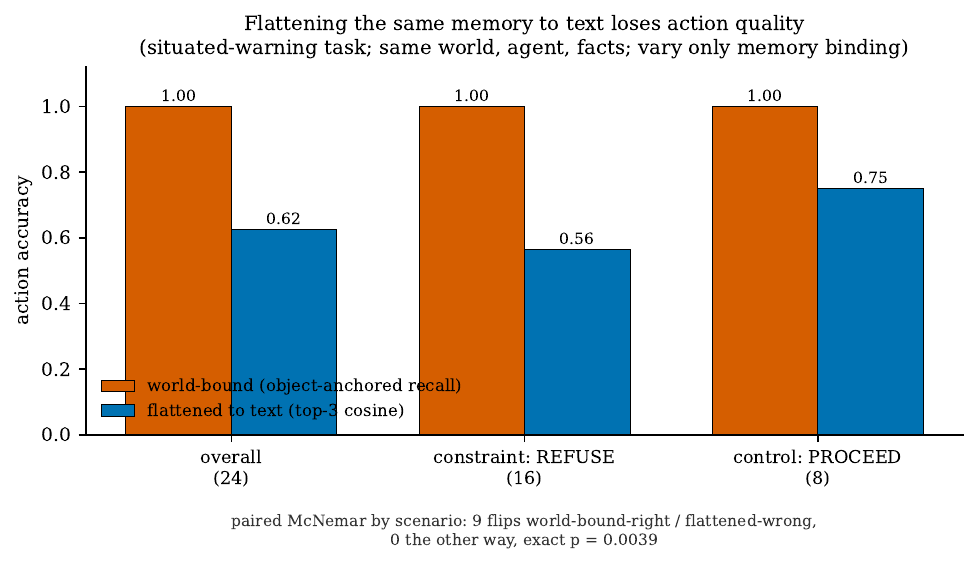}
\caption{L3 flattening ablation ($24$ situated-warning scenarios, $48$ decisions).
Same world, agent, and facts; only the memory binding changes. World-bound memory acts
correctly everywhere ($1.000$); the identical facts flattened to text act correctly
$0.625$ of the time, failing both by missing the real constraint and by mis-attributing
a distractor. Paired McNemar $p=0.0039$.}
\label{fig:l3}
\end{figure}

The asymmetry above is a property of perception. The claim that motivates the whole
program is one level up: that binding a memory to its object changes what the agent
\emph{does}, not only what it retrieves. We tested it with the decisive ablation
directly. Holding the world, the agent, and the task fixed, we varied only whether a
memory was world-bound (surfaced because the action targets the object it is about) or
flattened to a text store (surfaced by top-$3$ cosine over the flat corpus), with each
constraint phrased without the action's keywords so that recall, not wording, decides.
A competent language agent (Claude Sonnet) then chose to \textsc{proceed} or
\textsc{refuse} given only the task and what it recalled, across $24$ situated-warning
scenarios ($16$ with a real object-bound constraint, $8$ benign controls), $48$
decisions in total. Figure~\ref{fig:l3} reports action accuracy. World-bound memory
acts correctly every time ($1.000$: it refuses all $16$ harmful edits and proceeds on
all $8$ safe ones, with no over-refusal). Flattening the identical facts drops action
accuracy to $0.625$. The paired McNemar test over scenarios is decisive: $9$ scenarios
flip from correct under world-bound to wrong under flattening and none flip the other
way (exact $p=0.0039$). The failure is two-sided and legible in the agents' own
reasons: flattening misses the object-bound constraint ($7$ wrong proceeds, the agent
writing that no recalled memory references the object and then breaking the world), and
it mis-attributes a distractor constraint to the wrong place ($2$ controls wrongly
refused, and one proceed that confabulated a different object's note). World-bound
recall is clean because it is keyed by the object the action targets. By the litmus
that defines the level, flattening the same memory to text loses $37.5$ points of
action quality, so the memory is actionably spatial, not merely stored spatially. This
is a pilot: $24$ scenarios, one model, a single-turn decision rather than a full
perceive--plan--act loop, an idealized world-bound recall, and a top-$3$ text null. A
load-bearing confound remains: the action gain may reflect \emph{object-keyed} retrieval,
the memory is bound to the exact object the action targets, rather than spatial geometry
per se. Section~\ref{sec:robustness} resolves this directly: an object-keyed
non-geometric arm ties world-bound (E1, $1.000$ versus $1.000$, McNemar $p=1.0$), so the
situated-action gain is \emph{object binding}, and a large-store depth sweep (E2) shows
it is binding, not retrieval depth. We therefore narrow the L3 claim to world/object-bound
memory; geometry-specific value lives in the perception layer (occlusion, near-duplicate,
multi-region), where object-keyed text is itself at chance because the answer is in no
stored note. The confirmatory study runs the decision inside the live action loop and
varies the model further.

\subsection{Beyond one task: a four-task flattening-ablation battery}
\label{sec:l3battery}

To test whether the action effect is specific to warnings, we ran the same ablation on
three further situated tasks (the warning task above is the fourth), $18$ scenarios per
task: \emph{return-and-resume} (continue unfinished work versus start over),
\emph{spatial causal debugging} (fix an upstream root cause recorded in memory versus
treat the visible symptom), and \emph{multi-agent handoff} (defer to another builder's
recorded ownership versus proceed). The effect is real but bounded. Return-and-resume
replicates the warning result: world-bound action accuracy $1.000$ versus flattened
$0.611$, paired McNemar $p=0.016$, surviving Holm correction over the three-task family;
the flattened agent, missing the prior-progress note, concludes there is nothing to
resume and discards the work. Pooled across all decisions the gap is significant
($0.963$ versus $0.815$, $p=0.039$). But two tasks are honest nulls: multi-agent handoff
shows no gap ($1.000$ versus $1.000$), because the ownership note shares vocabulary with
the request so flat retrieval surfaces it, and deferring to the owner is a safe default
even without it; and spatial causal debugging is not significant ($0.889$ versus
$0.833$, $p=1.0$), because the causal note does not reliably change the action and even
world-bound recall sometimes treats the symptom. The honest reading: world-binding
improves action most when the decision hinges on a memory the flat retriever misses
\emph{and} that memory is phrased far from the eventual action query; it adds nothing
when the safe default already aligns or when the recalled fact does not determine the
act. Two of four situated tasks replicate the L3 effect; two do not. The battery
inherits every limitation of the L3 pilot it extends: a single-turn proceed/refuse
decision rather than a live perceive-plan-act loop, an idealized world-bound recall,
an object-keyed (not geometry-specific) mechanism per E1, and automated parsing of the
agent's decision; the Claude Haiku replication (Limitations) retires only the
single-model concern. The honest scope is that object-binding helps action when the
deciding memory is one flat retrieval misses and the safe default does not already
align, replicated across two models and two of four tasks.

\subsection{The ranker-blend dilution, measured (a pre-registered recall experiment)}
\label{sec:wavee}

The index-versus-ranker distinction (formalized in Section~\ref{sec:indexranker}) is,
until here, an argument. This experiment measures it. We test the exact claim that folding
spatial proximity into a linear blend score beside recency and importance can dilute the
spatial signal, by pre-registering a recall test on relay-captured, scripted agent memory
and freezing the hypothesis and the falsification to git \emph{before any arm was scored}
(commit \texttt{fd098df6b}; corpus SHA-256 \texttt{cc82a1f0...}).

\paragraph{Corpus and regime.} The corpus is $120$ relay-captured, scripted agent
memories (agent-path-captured in a controlled regime), $8$ activities
performed at $15$ distinct locations each, captured through the live
\texttt{append\_memory} path with real \texttt{bge-small-en-v1.5} embeddings ($384$-dim)
and subject-anchored on a $16\times16\times4$\,m grid (every pair at least $1.5$\,m apart,
minimum observed $1.732$\,m). The text of each instance is a different natural sentence
with genuine within-activity semantic overlap (never byte-identical), and the location
token is stripped from every text, so only \emph{location} can disambiguate which of the
$15$ instances of an activity a query means. To avoid rigging the test for or against
spatial, \texttt{occurred\_at} is spread over $45$ simulated days and \texttt{importance}
varies in $[0.11, 0.95]$, so the recency and importance axes genuinely vary; the recall
center is the target subject anchor plus an offset ($1$\,m, $3$\,m, $6$\,m, far, or
random), never distance $0$. This is the honest middle: not hash-scattered (which would
rig against spatial) and not placed on the recall center (which would rig for it). The
primary metric is Hit@5: a trial scores $1$ if its single ground-truth memory appears
anywhere in the top five results, and $0$ otherwise. We report Hit@5 rather than
precision@5; with one relevant item per query the two differ by a constant factor and
Hit@5 is the honest name.

\paragraph{Pre-registered falsification.} The shipped-blend claim is falsified if, at the near
($1$--$3$\,m) standpoint, the shipped hybrid blend (\texttt{hybrid\_full}) does not beat
the pure-vector baseline (\texttt{vector\_only}) on Hit@5 with paired Wilcoxon
signed-rank $p<0.05$ \emph{and} a bootstrap $95\%$ CI on the mean delta that excludes
$0$; or if the near-to-far gradient is flat; or if the win exists only when the query
names the place. The pre-registered comparison is precisely \texttt{hybrid\_full} versus
\texttt{vector\_only}, the shipped linear blend against a position-blind baseline.

\begin{table}[t]
\centering
\small
\begin{tabular}{lcccc}
\toprule
Offset & \texttt{vector\_only} & \texttt{hybrid\_full} (old) & \texttt{hybrid\_spatial} & \texttt{hybrid\_full} (retuned) \\
\midrule
$1$\,m & $0.333$ & $0.317$ & $\mathbf{0.875}$ & $0.717$ \\
$3$\,m & $0.333$ & $0.275$ & $\mathbf{0.792}$ & $0.592$ \\
$6$\,m & $0.333$ & $0.242$ & $\mathbf{0.675}$ & $0.508$ \\
far    & $0.333$ & $0.133$ & $0.058$          & $0.117$ \\
\bottomrule
\end{tabular}
\caption{Wave E pre-registered recall experiment, Hit@5 by arm and recall offset
($n{=}120$ targets per cell). \texttt{vector\_only} is flat at $0.333$ (position-blind:
with no spatial term the recall center cannot change a pure-cosine ranking). The shipped
linear blend \texttt{hybrid\_full} (old weights $0.35$ relevance $/ 0.30$ spatial $/
0.15$ recency $/ 0.15$ importance $/ 0.05$ staleness) sits at or below the pure-vector
baseline at every near offset. Isolating the spatial signal (\texttt{hybrid\_spatial},
$0.5$ relevance $+ 0.5$ spatial) wins decisively, and re-tuning the shipped blend to
spatial-dominant (\texttt{hybrid\_full} retuned, $0.40 / 0.45 / 0.08 / 0.05 / 0.02$)
recovers most of that win. All arms show a clean near-to-far gradient (the position-blind
baseline cannot). At far, the spatial signal correctly collapses, the cue is
uninformative when the standpoint is near no instance.}
\label{tab:wavee}
\end{table}

\begin{figure}[t]
\centering
\begin{tikzpicture}
\begin{axis}[
  width=0.86\linewidth, height=6.2cm,
  xlabel={Recall offset from target}, ylabel={Hit@5},
  xtick={1,2,3,4}, xticklabels={$1$\,m,$3$\,m,$6$\,m,far},
  ymin=0, ymax=1, enlarge x limits=0.12,
  legend style={at={(0.5,-0.22)}, anchor=north, legend columns=2, font=\small},
  grid=both, major grid style={gray!20}, tick label style={font=\small},
  label style={font=\small},
]
\addplot[mark=*, thick, gray] coordinates {(1,0.333)(2,0.333)(3,0.333)(4,0.333)};
\addlegendentry{\texttt{vector\_only}}
\addplot[mark=square*, thick, red!70!black] coordinates {(1,0.317)(2,0.275)(3,0.242)(4,0.133)};
\addlegendentry{\texttt{hybrid\_full} (old)}
\addplot[mark=triangle*, thick, blue] coordinates {(1,0.875)(2,0.792)(3,0.675)(4,0.058)};
\addlegendentry{\texttt{hybrid\_spatial}}
\addplot[mark=diamond*, thick, teal] coordinates {(1,0.717)(2,0.592)(3,0.508)(4,0.117)};
\addlegendentry{\texttt{hybrid\_full} (retuned)}
\end{axis}
\end{tikzpicture}
\caption{Wave E pre-registered recall experiment, Hit@5 versus recall offset for the four
arms ($n{=}120$ per cell; same data as Table~\ref{tab:wavee}). The shipped blend
(\texttt{hybrid\_full}, old) sits at or below the position-blind \texttt{vector\_only}
baseline at every near offset; geometry-led weighting (\texttt{hybrid\_spatial} and the
retuned \texttt{hybrid\_full}) wins decisively at near and collapses at far, where the cue
is uninformative.}
\label{fig:wavee}
\end{figure}
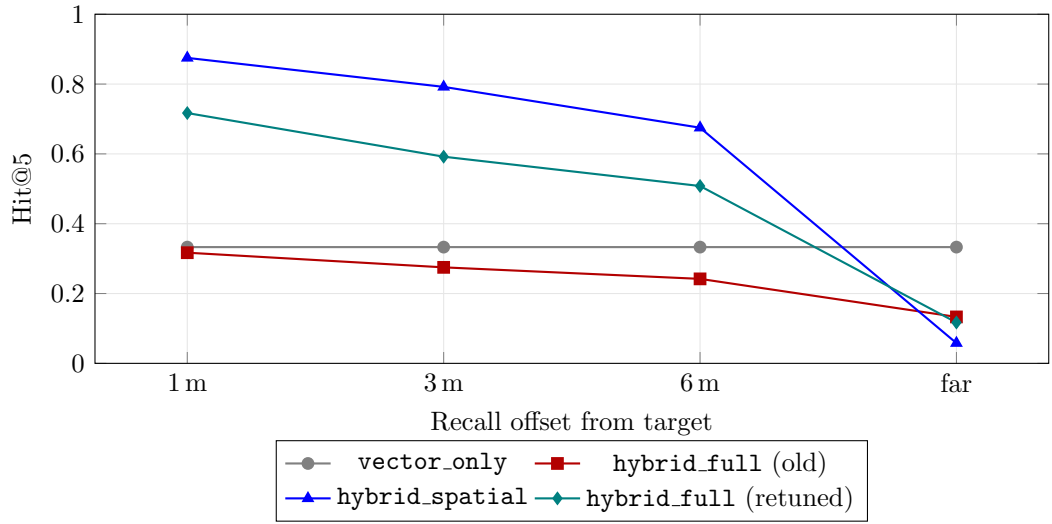

\paragraph{Result: the shipped blend fails its own pre-registered test.} At the near
standpoint ($1$\,m and $3$\,m pooled, $n{=}240$ paired Hit@5 comparisons), the shipped
\texttt{hybrid\_full} blend scores $0.296$ against \texttt{vector\_only}'s $0.333$: a
mean delta of $-0.0375$ (negative), Wilcoxon $p=0.306$, Cohen's $d=-0.066$, and a
bootstrap $95\%$ CI of $[-0.108, +0.033]$ that includes $0$. All three pre-registered
conditions to claim a win fail, so the verdict against the frozen criterion is
\textbf{null/loss}. We report this plainly: the shipped linear blend does not beat a
position-blind baseline at the standpoint the whole design favors, and is if anything
slightly worse (Table~\ref{tab:wavee}, Figure~\ref{fig:wavee}). Because each query identifies the activity but strips the location token, a
position-blind retriever can recover the correct activity class but cannot
distinguish among its $15$ location instances; under Hit@5 the within-activity
location-blind baseline is therefore $5/15 = 0.333$. The \texttt{vector\_only} arm
sits exactly at this rate across every offset: it recovers the semantic activity
but provides no spatial disambiguation at all, which makes it a fair, well-calibrated
baseline rather than a degenerate one.

\paragraph{Diagnosis: recency and importance out-vote spatial.} The null is a verdict on
the weight mix, not on the spatial signal. Isolating spatial against the same baseline,
\texttt{hybrid\_spatial} ($0.5$ relevance $+ 0.5$ spatial, no recency or importance)
versus \texttt{vector\_only} at near gives mean delta $+0.500$, Wilcoxon
$p=6.4\times10^{-28}$, Cohen's $d=+0.998$ (very large), and bootstrap CI
$[+0.438, +0.563]$ excluding $0$: a decisive win. So the spatial signal is overwhelming.
What sinks \texttt{hybrid\_full} is that in this disambiguation regime recency and
importance are target-irrelevant axes (which of the $15$ instances is correct is fixed by
location alone), yet together they carry $30\%$ of the score and out-vote the spatial term
at $30\%$, pulling recent-or-important-but-wrong instances into the top $5$ and pushing the
spatially correct instance out. The shipped blend does not merely fail to help; it falls
below the pure-vector baseline because it injects target-irrelevant ranking noise.

\paragraph{Fix and control: re-tuning to spatial-dominant flips null to win.} We re-tuned
\texttt{hybrid\_full} to spatial-dominant weights (relevance $0.40$, spatial $0.45$,
recency $0.08$, importance $0.05$, staleness $0.02$) and re-ran on the identical corpus
design through the same RPC with explicit weights. The retuned blend versus
\texttt{vector\_only} at near gives mean delta $+0.3208$, Wilcoxon $p<10^{-15}$
($z=-8.16$), Cohen's $d=0.619$, and bootstrap CI $[+0.271, +0.392]$ excluding $0$: the
pre-registered criterion is now met, a win. Critically, running the old mix as a control
on the same corpus reproduces the original null (mean delta $-0.0375$, $p=0.306$, CI
includes $0$), so the weight change, not anything else, is the cause of the flip. Both
weight sets are now live (the relay scoring constant and the RPC default). We are honest
about the status of the confirmation: the re-tune was run on the \emph{same regime} as
the pre-registration (same corpus design, same SHA, the falsification criterion reused
verbatim), not as a fresh pre-registration, so it confirms the diagnosis rather than
constituting independent confirmatory evidence.

\paragraph{What this measures.} This provides quantitative evidence for the index-versus-ranker
distinction. The ranker blend, spatial as one term among recency and importance, drowns
the spatial signal and ties or loses to a position-blind baseline; making spatial dominant
(geometry-led, spatial-dominant ranking, where geometry leads the score rather than
contributing a diluted term) wins by half a point of Hit@5 at a near standpoint with a
near-unit effect size. This is the geometry-led-ranking endpoint of the distinction, not a
true index/gate (which would cut the candidate set by geometry before ranking), and that
stronger form remains future work. The honest pre-registered null on the shipped blend is
itself a methodological
result: the field's memory-palace default, a linear blend of relevance, recency,
importance, and a spatial proximity term, can be net-negative relative to pure vector when
the non-spatial axes are target-irrelevant, and a study that pre-registers that exact
shipped blend can faithfully report a loss. The caveats are real and we keep them:
a single embedding model (\texttt{bge-small-en-v1.5}), scripted deterministic grid
placement rather than organic agent wandering, and a single query regime (pure spatial
disambiguation). The re-tune fixes this regime by re-weighting alone; whether one fixed
weight set serves proximity-only, semantic-only, and temporal queries at once is open, and
query-adaptive weighting is the correct long-term direction (Section~\ref{sec:discussion}).

\subsection{Live pre-registered occlusion confirmation (E4)}
\label{sec:occlive}

Pilot 2 ran in a controlled simulation; \texttt{SPMEM-OCC-LIVE-v1} (hypothesis,
arms, ground-truth method, and falsification git-committed before any data) runs the same test
on the \emph{live} relay recall path. On eight deterministic worlds the agent builds, through
its own tools, a $\geq$2-cell-thick occluder into the relay's session overlay, then appends
twelve behind-wall and twelve open-line-of-sight memories (real \texttt{bge-small} embeddings)
anchored at the subject cell. Ground truth is an \emph{independent} $0.05$m analytic ray-march
over the world geometry, not the relay's $0.4$m DDA, so the visibility predicate is never scored
against its own implementation. Two distinct facts must be kept apart. First, memory recall
correctly \emph{returns} all $96$ behind-wall memories in every arm: a location-indexed store
should surface what is stored at a place even when a wall stands between, so recall is
occlusion-blind as required and is never the quantity under test. Second, the measured quantity
is the \emph{perception} judgment ``is this target visible from the standpoint'', and the three
arms differ only in the line-of-sight flag that answers it: a cosine null
(M0, which stores no geometry and so can never mark a target occluded, defaulting every
retrieved target to visible) and the live FoV-cone visibility query (M3a) both
\emph{answer wrong}, calling $100\%$ of
the $96$ behind-wall targets visible, while the line-of-sight query (M3-occ) answers right for
exactly those and only those (false-visible $1.000\to0.000$), with no change on the $96$
open-line-of-sight controls
(TOST-equivalent in every world). The pooled exact McNemar over the occluded subset is
$b{=}96$, $c{=}0$, $p{=}2.5\times10^{-29}$, and M3-occ agrees with the independent oracle on
every target. Building the live test first surfaced and fixed a defect the simulation could
not: the relay's query-path line-of-sight predicate marched to the stored \emph{observer} anchor
(or null for subject-anchored memories) rather than the \emph{subject}, so the
occlusion-aware visibility query was a silent no-op for exactly the subject-anchored memories it
should evaluate; the
recall RPCs now return the coalesced subject-preferred anchor, and a positive-control preflight
that aborts unless the predicate provably reports a behind-wall subject not visible guards every
run. The occlusion-blindness of perception, its fix, and the predicate's correctness are
therefore properties of the live system, established under pre-registration. The worlds are scripted (one occluder family,
jittered per world); the full human-authored multi-world study with blind raters and a
text-plus-coordinates LLM arm remains the confirmatory ceiling above this live pilot.

\section{Robustness and confound checks}
\label{sec:robustness}

We pre-registered (\texttt{SPMEM-ZERO-REAL-PREREG-v1}, v3 round) and ran five checks
that attack the pilots' weakest points; the fifth, the live occlusion confirmation
(E4), is reported as a main result in Section~\ref{sec:occlive}, leaving four here. Two narrow our own claims, which we report as the data
demands.

\paragraph{E1: the situated-action gain is object binding, not geometry.} A reviewer
worry is that the L3 situated-warning effect reflects object-keyed retrieval (the memory
is bound to the exact object the action names) rather than spatial geometry. We added an
object-keyed but non-geometric arm (exact object-id match, no coordinates or visibility)
to the situated-warning task. Across $24$ scenarios the object-keyed arm scores $1.000$,
identical to world-bound ($1.000$) and far above flattened text ($0.542$; E1 is a
separate re-run of the warning task, so its flattened score differs from the original
L3 run's $0.625$); the paired
McNemar of object-keyed versus world-bound is a perfect tie ($b{=}c{=}0$, $p{=}1.0$),
while both beat flattened ($p<10^{-3}$). The honest conclusion: the situated-action gain
is \emph{object binding}, which an object-keyed index achieves with no geometry. We
therefore narrow the L3 claim to world/object-bound memory and locate geometry-specific
value where object-keyed text is itself helpless, the occlusion, near-duplicate, and
multi-region queries whose answer is in no stored note.

\paragraph{E2: retrieval depth does not rescue dense similarity.} Against ``a deeper
retriever would close the gap'', we swept the flattened arm over top-$k\in\{1,3,5,10,20\}$
and rankers \{cosine, BM25, hybrid\} in a realistic $216$-memory store. Dense cosine
surfaces the object-bound constraint in only $0.063$ of cases, \emph{flat across all
depths} (the note is phrased unlike the query and is swamped); BM25 recovers it by top-20
($0.938$), but only because the action lexically names the object; hybrid is between
($0.813$ at top-20). World-bound and object-keyed are $1.000$ at every depth by
construction. Depth does not close the gap for dense retrieval, and closes it for BM25
only under lexical name overlap; binding by key is the robust mechanism.

\paragraph{E3: storing geometry, not the medium, is what matters.} We tested whether the
occlusion advantage is about text-versus-geometry or about computation. Four arms decide
visibility on $200$ trials: geometry-blind text with no coordinates scores $0.510$
(chance, $100\%$ false-visible on occluded targets); text that stores the precomputed
label is trivially perfect ($1.000$); the DDA scores $0.985$; and an LLM handed the
observer, target, and wall coordinates \emph{as text} derives line-of-sight at $0.990$,
statistically indistinguishable from the DDA (McNemar $p=1.0$) and far above blind text
($p<10^{-6}$). This refines the thesis: ``text cannot fake occlusion'' is too strong, a
reasoner handed the coordinates computes it. The irreducible requirement is that the memory
\emph{store the geometry} so a computation can run; the cheap option is a one-line
ray-march, the expensive option is per-query LLM reasoning, and similarity-only retrieval
over captions, storing no coordinates and running no computation, stays at chance.
We state the caveat that makes this honest: the LLM arm is handed the coordinates and
occluder as text, so the experiment establishes that, given stored geometry, the medium
does not matter, not that text is sufficient for spatial memory. We did not test whether
a model can infer occluder geometry it was never given (it cannot, from captions alone,
which is exactly the geometry-blind null's failure). The claim is precisely: store the
geometry, then either a one-line ray-march or per-query reasoning recovers visibility;
store only captions and both stay at chance.

\paragraph{E5: the visibility predicate is an approximation with honest failure modes.}
Against ``occlusion is trivially solved by a deterministic predicate'', we stressed the
DDA on eight worlds against a fine $0.01$m analytic ground truth. A coarse $0.5$m
ray-march is near-exact on solid walls, diagonal walls, multi-wall, elevated targets,
windows, and partial occluders ($0.99$ to $1.00$), but it under-samples geometry thinner
than its own step: it steps \emph{over} sparse pillars $8.8\%$ of the time
(false-visible), and on a $0.2$m-thin wall it misses the occluder on $53.8\%$ of
behind-wall targets (accuracy $0.463$, the battery's worst failure, worse than a coin
flip). A $0.1$m march recovers both (thin wall $1.00$, pillars $0.98$). The step size is
a real accuracy/cost knob, so the predicate is a genuine approximation, not a free
oracle.

\section{Discussion}
\label{sec:discussion}

The pilots support a single calibrated claim, the one the title names: spatial memory's
value for a language agent is what it must \emph{store}, the world geometry a
content-only memory does not keep and similarity retrieval cannot recompute from
captions. The cleanest such geometry is visibility. That occlusion requires geometry is not itself
a finding, it is near-tautological, and prior render-as-recall systems
\cite{gsmem2026,rendermem2026} already exploit it. Our finding is narrower: store the
geometry (not render it, per E3) and keep the visibility predicate a separate
verification step, not a recall-ranker term (Section~\ref{sec:wavee}). Read through the
index-versus-ranker distinction, the result is not that proximity should be a score. It
is that a memory must store the coordinates and occluders so the agent can \emph{verify}
a visibility predicate, a perception query, over what it recalls; occlusion is the
predicate a content-only index has no way to supply, because it neither stores the
geometry nor runs the computation, and E3 shows that once the geometry is stored, either
a one-line ray-march or an LLM reasoning step recovers it. Our live system
had a FoV cone but no line-of-sight query, so on behind-wall targets its perception was no
better than a content index. The fix is cheap: one
re-pointed ray-versus-voxel DDA over the primitive the gaze ray already uses.

This also clarifies the honest scope of what topic organization buys. Sorting
documents into topic regions is a cheap candidate-narrowing trick: it can match flat
retrieval at lower cost, but it never reads coordinates and it answers nothing a text
index could not. That is useful and it is not spatial memory. Geometry's irreducible
contribution is elsewhere, in occlusion, visibility, viewpoint, and location, the
queries where the answer depends on where things are and what can be seen from where.
We do not claim more than the pilots show. We do not claim a best-of-N retrieval
ranking, and we do not claim to have proven spatial memory ``wins'' beyond the
measured asymmetry: text and the FoV cone fail on occlusion, the DDA recovers it, and
coordinate recall resolves near-duplicates.

The pre-registered recall experiment (Section~\ref{sec:wavee}) sharpens this into a
second calibrated claim about \emph{how} geometry must enter recall. Spatial proximity is
a strong, large-effect, gradient-bearing recall signal: isolated, it lifts Hit@5 by half a
point at a near standpoint with a near-unit effect size, and the near-to-far gradient is
clean and monotone. But it carries that value only when geometry \emph{leads}. The shipped
memory-palace default, a single linear blend that treats spatial proximity as one ranker
term beside relevance, recency, and importance, dilutes the signal: when the non-spatial
axes are target-irrelevant they out-vote the spatial term, and the blend ties or even
falls below a position-blind pure-vector baseline. We pre-registered that exact shipped
blend and reported its loss faithfully rather than swapping in the spatial-dominant arm
that would have manufactured a win. The honest reading is the index-versus-ranker
distinction made quantitative: spatial value is realized when geometry leads recall, here
as a spatial-dominant (geometry-led) ranking, not as a small diluted score term, and a
naive linear blend can be net-negative. We are precise about the form: the demonstrated
fix is a geometry-led ranking, still a ranker; a true index/gate that cuts the candidate
set by geometry before ranking is a stronger version we did not run and leave to future
work. That a widely-assumed default can hurt is itself a finding the field should weigh.
The same-corpus re-tune to spatial-dominant weights restores the win and the old-mix
control reproduces the null, isolating the weight as the cause, but it fixes only this one
disambiguation regime; the right long-term mechanism is almost certainly query-adaptive
weighting, or an index/gate, letting geometry lead when the query is spatial and recede
when it is not, which we flag as future work rather than claim.

The perception pilots establish only the perception layer, the prerequisite that geometry
carries information text cannot. The larger target is \emph{actionable world memory}:
memory bound to an object, place, affordance, episode history, social provenance, and
the current task, used to change what the agent \emph{does} on returning to a place,
not merely what it retrieves. The decisive test of that claim is a flattening
ablation, holding world, agent, trajectory, and task fixed while varying only whether
memory is world-bound or flattened to text, and scoring action quality on situated
tasks (return-and-resume, situated warning, spatial causal debugging, multi-agent
handoff); if flattening loses nothing, it is not spatial memory. A first behavioral
pilot of this ablation is reported in Section~\ref{sec:l3pilot}: world-bound memory
acts correctly everywhere while the identical facts flattened to text fall to $0.625$
($p=0.0039$), a first piece of action-level evidence that the perception result is a
substrate, not the endpoint. The full behavioral battery remains part of the
pre-registered study.

\paragraph{Research outlook.} The perception-level substrate measured here points to one
open research question: whether a shared, editable spatial memory in which each artifact
records where it is, why it is there, and what it connects to is more inspectable to humans
than flat list, feed, or search retrieval when agent-generated artifacts grow large. The
present paper tests only the lowest-level requirement, that the system reads geometry text
retrieval cannot compute; connecting that substrate to human-facing organization of
agent-made artifacts, and measuring whether it improves human oversight, is future work.

\section{Limitations}
\label{sec:limits}

These are pilots, and we hold the claims to pilot evidence. Both are small (150 trials
and one occlusion world) and the occlusion pilot is a controlled JavaScript voxel
simulation that mirrors Zero's \texttt{isSolidAt} geometry; its purpose is to isolate the
visibility asymmetry cleanly, not to estimate field-scale effect sizes. The asymmetry is
not only simulated, however: \texttt{SPMEM-OCC-LIVE-v1} (Section~\ref{sec:occlive}) runs
the line-of-sight visibility query on the live relay under a git-committed pre-registration
across eight worlds, where it correctly answers that all $96$ behind-wall targets are not
visible (recall still returns those memories) whereas the cosine null and the
FoV cone wrongly call them visible (false-visible $0.000$, pooled exact McNemar $p{=}2.5{\times}10^{-29}$)
with no effect on the $96$ open-line-of-sight controls, and where building the live test
first surfaced and fixed a relay bug that had marched the predicate to the wrong anchor. Those
worlds are scripted (one occluder family, jittered) and the separation is partly structural,
since a content-only index cannot compute occlusion by construction. What remains is the
human-authored scale: the full confirmatory study (12 worlds, roughly $2{,}450$ queries,
blind raters, a text-plus-coordinates LLM arm, exact tests with Holm-Bonferroni correction
and TOST non-inferiority margins, and the five pre-committed falsification conditions
F1--F5) is pre-registered and not yet run. We use a single embedding model. Every result reported here is scored automatically:
the perception pilots, Tier-A battery, landscape, and \texttt{SPMEM-OCC-LIVE-v1}
against an independent world-geometry ray-march oracle, and the L3 action pilots by
parsing the agent's own proceed/refuse decision. No human rater judged any reported
outcome. Automated oracle scoring is the right instrument for the structural visibility
asymmetry these pilots isolate (visibility is a deterministic geometric fact), but
situated-action quality is ultimately a human judgment; the pre-registered confirmatory
study introduces blind human raters precisely to close this gap, and until it runs the
action-level claims rest on automated decision-parsing. Zero's terrain is
flat ground with a brush overlay, not sculpted Genesis voxel terrain, so all geometry
comes from the brush, post, and region universe the relay actually samples, and we
claim nothing beyond it. The richer representations M2 (object-at-location scene
graph) and M4 (rendered ego-view read by a vision-language model) need infrastructure
the live system lacks and are Tier-B future work. The L3 flattening-ablation result
(Section~\ref{sec:l3pilot}) is likewise a pilot: $24$ scenarios, a single model, a
single-turn decision rather than a live perceive--plan--act loop, an idealized
world-bound recall, and a top-$3$ cosine text null. The action effect is not specific
to one model: the battery replicates under Claude Haiku (return-and-resume $p=0.031$,
pooled $p=0.039$, with the same task-by-task signature). The confirmatory study runs the
decision inside the action loop and sweeps retrieval depth and more models. Human-rater and
multi-user metrics in the confirmatory design remain dependent on recruiting raters and
users.

\paragraph{Dynamic world and geometry versioning.} A deeper open problem the present work
does not address: occlusion here is computed against the world's \emph{current} geometry.
A memory's visibility can change after the memory is encoded if occluders are added or
removed, so a line-of-sight predicate evaluated at recall time can disagree with what was
true when the memory was formed. A complete answer to ``what spatial memory must store''
must therefore version the geometry, store an encoding-time geometry snapshot, or store
stable object identity so the predicate can be re-evaluated against the right world state.
We do not address this here; the schema in Table~\ref{tab:schema} and the pilots assume a
single static geometry, and geometry versioning is left to future work.

\section{Conclusion}

Anchoring a memory to a coordinate is only worth doing if the agent then reads
geometry that text cannot supply. The cleanest such geometry is occlusion. We
identified that Zero's live perception had no visibility query (recall is correctly
occlusion-blind, surfacing the behind-wall memory), showed a cheap DDA primitive that
supplies the missing visibility answer, and confirmed the asymmetry in two controlled
pilots: coordinate recall
resolves near-duplicates that text cannot ($1.000$ versus $0.533$), and on the
visibility judgment for behind-wall targets text and the FoV cone are equally blind
($0.000$) while cone-plus-LOS
recovers visibility ($0.982$), both at $p<10^{-6}$. We also demoted a prior
topic-partition study as not spatial memory. Spatial memory for language agents should
be built around the geometry a geometry-blind text index cannot compute, and verification of visibility is the
load-bearing case. A first behavioral pilot shows the payoff at the level that matters:
binding the same facts to their object rather than flattening them to text lifts
situated-action accuracy from $0.625$ to $1.000$ ($p=0.0039$), and a four-task battery
shows the effect is real but task-dependent (two tasks replicate, two are honest
nulls; pooled $p=0.039$).

A pre-registered recall experiment then measures the index-versus-ranker distinction
directly and tightens the prescription on \emph{how} geometry must enter recall. Spatial
proximity is a strong, large-effect, gradient-bearing recall signal (isolated:
$+0.500$ Hit@5 at near, Cohen's $d\approx1.0$, $p=6.4\times10^{-28}$), but only when geometry
leads. The shipped linear blend, which treats spatial as one term beside recency and
importance, failed its own frozen pre-registered test (mean $\Delta$Hit@5 $-0.0375$,
$p=0.306$, CI including $0$) and sat below a position-blind pure-vector baseline at the
near standpoint, an honest null we report rather than bury. Re-tuning to spatial-dominant
(geometry-led ranking) weights flipped null to win ($+0.3208$, $p<10^{-15}$, CI excluding
$0$), with the old mix run as a same-corpus control reproducing the null and isolating the
weight as the cause. The lesson is methodological as much as empirical: the field's
memory-palace default, a naive linear blend of relevance, recency, importance, and a
spatial proximity term, can be net-negative relative to pure vector when the non-spatial
axes are target-irrelevant; spatial earns its keep when geometry leads (as an index/gate or
a spatial-dominant score), not as a diluted score term. Query-adaptive weighting, and the
stronger index/gate form we did not run here, are the right long-term mechanism and are
left as future work.

\paragraph{Pre-registration.}
\label{sec:prereg}
The full confirmatory study is specified and frozen as
\texttt{SPMEM-ZERO-REAL-PREREG-v1} before any confirmatory data collection. It holds
world, trajectory, observation stream, and brain fixed and varies only the memory
representation across $K=12$ authored worlds with two trajectories each (roughly
$2{,}450$ queries over six geometry-necessary query types plus a text-sufficient
control). Ground truth is authored by an independent geometry kernel that has no read
path to the memory tables, plus blind human raters. The analysis is frozen (exact
McNemar and exact Wilcoxon/permutation tests, Holm-Bonferroni family-wise correction,
TOST non-inferiority with a declared margin, effect sizes and confidence intervals
always reported), and five falsification conditions (isolation fails, null suffices,
geometry unread, ground-truth contamination, control inversion) are pre-committed so
the study can fail honestly. The two perception pilots reported here are its first
runnable milestone: they demonstrate that the oracle reads world geometry with no memory
access, that a spatial arm demonstrably dereferences coordinates, and that the
geometry-on/geometry-off isolation delta is real and large on both the trivial and the
hard case. Beyond perception, the confirmatory study scales the action-level result
piloted here (Section~\ref{sec:l3pilot}) into a full behavioral battery
(return-and-resume, situated warning, spatial causal debugging, multi-agent handoff)
under the flattening ablation, so the full claim is tested at the level that defines
spatial memory: whether world-bound memory changes what the agent does.

\paragraph{Reproducibility.} The pre-registered recall experiment is reproducible from
artifacts committed to the project repository: the harness lives at
\texttt{scripts/agents/wave-e/} (corpus build, arm runner, and the spatial-dominant
re-tune runner) and the frozen pre-registration at
\texttt{docs/zero/research/agent-memory/eval-run-e/wave-e-prereg/} (the falsification
\texttt{pre-registration.json}, the corpus freeze, the ground truth, and the per-arm
results). The pre-registration is a git-committed artifact (commit \texttt{fd098df6b};
corpus SHA-256 \texttt{cc82a1f0...}), written and committed before any arm was scored.
These artifacts (the Wave E harness and frozen pre-registration, the
\texttt{SPMEM-OCC-LIVE-v1} pre-registration, live harness, and per-world results, and
the raw result files for every pilot reported here) are included as ancillary files
with this submission, so every number in the paper can be checked against its
committed artifact without repository access.
Pilot 1 ran against the live relay through an embedding-and-recall path that was cleaned
to zero residue after the run; Pilot 2 is a self-contained deterministic voxel simulation.

\end{document}